%% file: main.tex
\documentclass{article}
\pdfoutput=1

\PassOptionsToPackage{round,sort}{natbib}

\usepackage[preprint]{neurips_2020}

\usepackage[utf8]{inputenc} 
\usepackage[T1]{fontenc}    
\usepackage{url}            
\usepackage{booktabs}       
\usepackage{amsfonts}       
\usepackage{nicefrac}       
\usepackage{microtype}      

\usepackage[colorlinks=true,%
citecolor=blue,%
filecolor=blue,%
linkcolor=blue,%
urlcolor=blue]{hyperref}

\input{our_files/preamble.tex}

\title{Differentiable PAC-Bayes Objectives with Partially Aggregated Neural Networks}

\author{%
  Felix Biggs \\
  Centre for Artificial Intelligence \\
  University College London \\
  United Kingdom \\
  \url{fbiggs@cs.ucl.ac.uk}
  \And
  Benjamin Guedj \\
  Centre for Artificial Intelligence \\
  Inria and University College London \\
  United Kingdom \\
  \url{b.guedj@ucl.ac.uk}
}

\begin{document}

\maketitle

\begin{abstract}
\input{our_files/abstract.tex}
\end{abstract}

\renewcommand\vec{\mathbf}

\input{our_files/section_introduction.tex}
\input{our_files/section_background.tex}
\input{our_files/section_aggregates.tex}
\input{our_files/section_sign_output.tex}
\input{our_files/section_objective_bound.tex}
\input{our_files/section_experiments.tex}
\input{our_files/section_discussion.tex}




\bibliographystyle{plainnat}
\bibliography{bibliography}
\setcitestyle{authoryear,open={(},close={)}}

\clearpage
\appendix
\input{our_files/appendix_experiments.tex}

\end{document}

%% file: our_files/abstract.tex
We make three related contributions motivated by the challenge of training
stochastic neural networks, particularly in a PAC-Bayesian setting: (1) we show
how averaging over an ensemble of stochastic neural networks enables a new class
of \emph{partially-aggregated} estimators; (2) we show that these lead to
provably lower-variance gradient estimates for non-differentiable signed-output
networks; (3) we reformulate a PAC-Bayesian bound for these networks to derive a
directly optimisable, differentiable objective and a generalisation guarantee,
without using a surrogate loss or loosening the bound. This bound is twice as
tight as that of \cite{NIPS2019_8911} on a similar network type. We show
empirically that these innovations make training easier and lead to competitive
guarantees.

%% file: our_files/section_introduction.tex
\section{Introduction}

The use of stochastic neural networks has become widespread in the PAC-Bayesian
and Bayesian deep learning \citep{pmlr-v37-blundell15} literature as a way to
quantify predictive uncertainty. PAC-Bayesian theorems generally bound the
expected loss of \emph{stochastic} estimators, so it has proven easier to obtain
non-vacuous numerical guarantees on generalisation in such networks, beginning
with \citet{NIPS2001_1968}, picked up by \citet{dziugaite2017computing} in the
modern ``deep'' era, and continued by others including
\citet{zhou2018nonvacuous}.

These bounds can often be straightforwardly adapted to aggregates or averages of
estimators (as in, for example, \citet{germainPACBayesianLearningLinear2009}),
but averages over deep stochastic networks are generally intractable.
\cite{NIPS2019_8911} successfully aggregated networks with sign activation
functions \(\in \{+1, -1\}\) and a non-standard tree structure, but this incurred
an exponential KL divergence penalty, and a heavy computational cost that meant
in practice they often resorted to a Monte Carlo estimate.

Our first innovation, not specific to PAC-Bayes, is to find a compromise by
defining new ``partially-aggregated'' Monte Carlo estimators for the average
output and gradients of general stochastic networks (\Cref{aggregates}). In the
case of non-differentiable ``signed-output'' networks (with a dense final layer
and sign activation, the rest of the network having arbitrarily complex
structure) we prove that these have lower variance than REINFORCE
\citep{williams1992simple} and a naive Monte Carlo forward pass
(\Cref{sign_output}). This framework additionally leads to a similar but simpler
training method for sign-activation neural networks than \citet{NIPS2019_8911}.

Next, we observe that when training \emph{neural networks} in the PAC-Bayesian
setting, the objective used is generally somewhat divorced from the bound on
misclassification loss itself, because non-differentiability leads to
difficulties with direct optimisation. For example,
\citet{dziugaite2017computing} use an objective with both a surrogate loss
function and a different dependence on the KL from their bound, and
\citet{NIPS2019_8911} use a different loss function which leads to a bound on
the misclassification loss which is a factor of two worse. This contrasts with
the situation for other estimator classes, where the bound may directly lead to
a differentiable objective (for example in
\citet{germainPACBayesianLearningLinear2009}).

Motivated by this, we show (\Cref{bounds}) that a straightforward adaptation of
a bound from \citet{catoniPacBayesianSupervisedClassification2007} to our
signed-output networks, in combination with aggregation, yields straightforward
and directly differentiable objectives: one that fixes a regularisation
parameter, and another that tunes it automatically through the bound. This
closes the previous gap between bounds and objectives in neural networks. We
demonstrate that training these objectives in combination with our partial
aggregation estimators leads to competitive experimental generalisation
guarantees (\Cref{experiments}). A brief discussion follows in
\Cref{discussion}.

%% file: our_files/section_background.tex
\section{Background}\label{background}

We begin here by setting out our notation and the requisite background.

The standard statistical learning perspective considers parameterised functions,
\(\{f_{\theta}: \mathcal{X} \to \mathcal{Y} | \theta \in \Theta\}\), in a specific form, choosing
\(\mathcal{X} \subset \mathbb{R}^{d_0}\) and an arbitrary output space \(\mathcal{Y}\) which
could be for example \(\{-1, +1\}\) or \(\mathbb{R}\). In general, we wish find functions
minimizing the out-of-sample expected risk,
\(R(f) = \mathbb{E}_{(\vec{x}, y) \sim \mathcal{D}} \ell(f(\vec{x}), y)\), for some
loss function \(\ell\), for example the 0-1 misclassification loss for
classification, \(\ell_{0-1}(y, y') = \mathbf{1}\{y \neq y'\}\), or the binary linear
loss, \(\ell_{\text{lin}}(y, y') = \frac{1}{2}(1 - yy')\), with
\(\mathcal{Y} = \{+1, -1\}\). These must be chosen based on an i.i.d. sample
\(S = \{(\vec{x}_i, y_i)\}_{i=1}^m \sim \mathcal{D}^m\) from the data distribution \(\mathcal{D}\),
using the surrogate of in-sample empirical risk,
\(R_S(f) = \frac{1}{m} \sum_{i=1}^m \ell(f(\vec{x}_i), y_i)\). We denote the expected
and empirical risks under the misclassification and linear losses respectively
\(R^{0-1}, R^{\text{lin}}, R_S^{0-1}\) and \(R_S^{\text{lin}}\).

In this paper we consider learning a distribution (PAC-Bayesian Posterior),
\(Q\), over the parameters \(\theta\). PAC-Bayesian theorems then provide bounds on
the expected generalization risk of stochastic classifiers, where every
prediction is made using a newly sampled function from our posterior,
\(f_{\theta}, \, \theta \sim Q\).

We also consider averaging the above to obtain \(Q\)-aggregated prediction
functions,
\[ F_{Q}(\vec{x}) := \E_{\theta \sim Q} f_{\theta}(\vec{x}).\]

In the case of a convex loss function, Jensen's inequality lower bounds the risk
of the stochastic function by its \(Q\)-aggregate:
\(\ell(F_Q(x), y) \leq \mathbb{E}_{f \sim Q} \ell(f(x), y)\). This is an equality for the linear
loss, which we will exploit to obtain an easier PAC-Bayesian optimisation
objective in Section \ref{bounds}.

\subsection{Analytic Q-Aggregates for Signed Linear Functions}\label{aggregated_signed_linear}

\(Q\)-aggregate predictors are analytically tractable for ``signed-output''
functions\footnote{Here the sign function and ``signed'' functions have outputs
\(\in \{+1, -1\}\), as the terminology ``binary'', as used in
\cite{NIPS2019_8911}, suggests to us too strongly an output \(\in \{0, 1\}\).} of
the form \(f_{w}(\vec{x}) = \sign(\vec{w} \cdot \vec{x})\) under a normal
distribution, \(Q(w) = N(\vec{\mu}, \mathbb{I})\), as first considered in a
PAC-Bayesian context for binary classification by
\cite{germainPACBayesianLearningLinear2009}, obtaining an differentiable
objective similar to the SVM. Provided \(\vec{x} \neq \vec{0}\):

\begin{equation}\label{erf_aggregate}
F_Q(\vec{x}) := \E_{\vec{w} \sim N(\vec{\mu}, \mathbb{I})} \sign(\vec{w} \cdot \vec{x}) = \erf\left( \frac{\vec{\mu} \cdot \vec{x}}{\sqrt{2} \|\vec{x}\|} \right).
\end{equation}

In Section \ref{sign_output} we will consider aggregating signed output
(\(f(x) \in \{+1, -1\}\)) functions of a more general form.

\subsection{Monte Carlo Estimators for More Complex Q-Aggregates}

The framework of \(Q\)-aggregates can be extended to less tractable cases (for
example, with \(f_\theta\) a stochastic or ``Bayesian'' neural network, see
\cite{pmlr-v37-blundell15}) through a simple Monte Carlo approximation:

\begin{equation}\label{basic_monte_carlo}
F_{Q}(\vec{x}) = \E_{\theta \sim Q} f_{\theta}(\vec{x}) \approx \frac{1}{T} \sum_{t=1}^T f_{\theta^{t}}(\vec{x}) := \hat{F}_{Q}(\vec{x}).
\end{equation}

If we go on to parameterize our posterior \(Q\) by \(\phi\) as \(Q_\phi\) and wish to
obtain gradients without a closed form for
\(F_{Q_{\phi}}(\vec{x}) = \E_{\theta \sim Q_\phi} f_\theta(\vec{x})\), there are two possibilities.
One is REINFORCE \citep{williams1992simple}, which requires only a
differentiable density function, \(q_\phi(\theta)\) and makes a Monte Carlo
approximation to the left hand side of the identity
\(\nabla_\phi \E_{\theta \sim q_\phi} f_\theta(\vec{x}) = \E_{\theta \sim q_\phi} [f_\theta(\vec{x}) \nabla_\phi \log q_\phi(\theta)]\).

The other is the pathwise estimator \citep[used in the context of neural
networks by][amongst others]{kingma2013auto}, which additionally requires that
\(f_\theta(\vec{x})\) be differentiable w.r.t. \(\theta\), and that the probability
distribution chosen has a standardization function, \(S_\phi\), which removes the
\(\phi\) dependence: for example, \(S_{\mu, \sigma}(X) = (X - \mu)/\sigma\) for a normal
distribution. If this exists, the right hand side of
\(\nabla_\phi \E_{\theta \sim q_\phi} f_\theta(\vec{x}) = \E_{\epsilon \sim p} \nabla_\phi f_{S_\phi^{-1}(\epsilon)}(\vec{x})\)
generally yields lower-variance estimates than REINFORCE \cite[see][for a modern
survey]{mohamed2019monte}.

The variance introduced by REINFORCE can make it difficult to train neural
networks when the pathwise estimator is not available, for example when
non-differentiable activation functions are used. Below we find a compromise
between the analytically closed form of \eqref{erf_aggregate} and the above
estimator that enables us to make differentiable certain classes of network and
extend the pathwise estimator where otherwise it could not be used. Through this
we are able to stably train a new class of network.

\subsection{PAC-Bayesian Approach}

We use PAC-Bayes in this paper to obtain generalisation guarantees and
theoretically-motivated training methods. The primary bound utilised is based on
the following theorem, valid for a loss taking values in \([0, 1]\):

\begin{theorem} \citep[Theorem
1.2.6]{catoniPacBayesianSupervisedClassification2007} Given \(P\) on
\(\mathcal{F}\) and \(\alpha > 1\), for all \(Q\) on \(\mathcal{F}\) with probability
at least \(1 - \delta\) over \(S \sim \mathcal{D}^m\),
\[ \mathbb{E}_{f \sim Q}R(f) \leq \inf_{\lambda > 1} \Phi_{\lambda/m}^{-1} \left[ \mathbb{E}_{f \sim Q}R_S(f) + \frac{\alpha}{\lambda} \left( \KL{Q|P} - \log \delta + 2\log \left(\frac{\log \alpha^{2}\lambda}{\log{\alpha}}\right) \right) \right] \]
with \(\Phi_\gamma^{-1}(t) = \frac{1 - \exp(-\gamma t)}{1 - \exp(-\gamma)}\).\label{catonizhou}
\end{theorem}

This slightly opaque formulation \citep[used previously
by][]{zhou2018nonvacuous} gives essentially identical results when
\(\text{KL}/m\) is large to the better-known ``small-kl'' bounds originated by
\citet{Langford01boundsfor,seeger2001improved}. It is chosen because it leads to objectives that
are \emph{linear} in the empirical loss and KL divergence, like

\begin{equation} \label{basicobjectiveform}
\mathbb{E}_{f \sim Q}R_S(f) + \frac{\KL{Q|P}}{\lambda}.
\end{equation}

This objective is minimised by a Gibbs Posterior and is closely related to the
evidence lower bound (ELBO) usually optimised by Bayesian Neural Networks
\citep{pmlr-v37-blundell15}. Such a connection has been noted throughout the
PAC-Bayesian literature; we refer the reader to \cite{NIPS2016_6569} or
\cite{knoblauchGeneralizedVariationalInference2019} for a formalised treatment.

%% file: our_files/section_aggregates.tex
\section{Partial Aggregation of Stochastic Neural Networks}\label{aggregates}

Here we consider a reformulation of \(Q\)-aggregation for a general class of
neural networks that leads to a different Monte Carlo estimator for their
outputs and gradients. These networks are those with a dense final layer, taking
the form

\[f_\theta(\vec{x}) = A(\vec{w} \cdot \vec{\eta}_{\thetanotw}(\vec{x})) \]

with \(\theta = \operatorname{vec}(\vec{w}, \thetanotw) \in \Theta\), \(\vec{w} \in \mathbb{R}^d\),
\(\vec{\eta}_{\thetanotw}: \mathcal{X} \to \mathcal{A}^d \subseteq \mathbb{R}^d\) and
\(A: \mathbb{R} \to \mathcal{Y}\). \(\thetanotw \in \Theta^{\notw}\) is the parameter set
excluding \(\vec{w}\), for the non-final layers of the network. For simplicity
we have used a one-dimensional output but we note that the formulation and below
derivations trivially extend to a vector-valued function with elementwise
activations. We require the distribution over parameters to factorise like
\(Q(\theta) = Q^{\vec{w}}(\vec{w}) Q^{\notw}(\thetanotw)\), which is consistent with
the literature.

We recover a similar functional form to that considered in
\Cref{aggregated_signed_linear} by rewriting the function as
\(A(\vec{w} \cdot \vec{a})\) with \(\vec{a} \in \mathcal{A}^d\) the hidden-layer
activations. The ``aggregated'' activation function on the final layer, defined
as \(I(\vec{a}) := \int A(\vec{w} \cdot \vec{a}) \D{Q^{\vec{w}}(\vec{w})}\), may then be
analytically tractable, as in \eqref{erf_aggregate}, where with
\(\vec{w} \sim N(\vec{\mu}, \mathbb{I})\) and a sign final activation,
\(I(\vec{a}) = \erf\left( \frac{\vec{\mu} \cdot \vec{a}}{\sqrt{2} \|\vec{a}\|} \right)\).

With these definitions we can write the \(Q\)-aggregate in terms of the
conditional distribution on the activations, \(\vec{a}\), which is a
push-forward measure,
\(\tilde{Q}^{\notw}(\vec{a}|\vec{x}) := (\vec{\eta}_{(\cdot)}(\vec{x})) \circ Q^{\notw}\),
(i.e. the distribution of \(\vec{\eta}_{\thetanotw}(\vec{x}) | \vec{x}\), with
\(\thetanotw \sim Q^{\notw}\)) as

\begin{align*}
F_Q(\vec{x}) := \E_{\theta \sim Q} [f_\theta(\vec{x})]
&= \int_{\theta^{\notw}} \left[ \int_{\mathbb{R}^d}  A(\vec{w} \cdot \vec{\eta}_{\thetanotw}(\vec{x})) \D{Q^{\vec{w}}(\vec{w})} \right] \D{Q^{\notw}(\thetanotw)} \\
&= \int_{\theta^{\notw}} I(\vec{\eta}_{\thetanotw}(\vec{x})) \D{Q^{\notw}(\thetanotw)} \\
&= \int_{\mathcal{A}^d} I(\vec{a}) \D{\{(\vec{\eta}_{(\cdot)}(\vec{x}))_\sharp Q^{\notw}\}(\vec{a})} \\
&=: \int_{\mathcal{A}^d} I(\vec{a}) \D{\tilde{Q}^{\notw}(\vec{a}|\vec{x})}. \\
\end{align*}

In most cases, the final integral cannot be calculated exactly or involves a
large summation, so we resort to a Monte Carlo estimate, for each \(\vec{x}\)
drawing \(T\) samples of the activations,
\(\{\vec{a}^t\}_{t=1}^T \sim \tilde{Q}^{\notw}(\vec{a}|\vec{x})\) to obtain the
``partially-aggregated'' estimator

\begin{equation}\label{aggregated_estimator}
F_Q(\vec{x}) = \int_{\mathcal{A}^{d}} I(\vec{a}) \D{\tilde{Q}^{\notw}(\vec{a}|\vec{x})} \approx \frac{1}{T} \sum_{t=1}^T I(\vec{a}^{t}) = \hat{F}_Q^{*}(\vec{x}).
\end{equation}

This is quite similar to the original estimator from \eqref{basic_monte_carlo},
but in fact the aggregation of the final layer may significantly reduce the
variance of the outputs and also make better gradient estimates possible, as we
will show below in the case of ``signed''-output networks.

\subsection{Single Hidden Layer}

For clarity (and to introduce notation used in \Cref{all_sign}) we will briefly
consider the case of a neural network with one hidden layer,
\( f_\theta(\vec{x}) = A_2(\vec{w}_2 \cdot \vec{A}_1(W_1 \vec{x})) \), with parameters
\(\theta = \operatorname{vec}(\vec{w}_2, W_1)\), \(W_1 \in \mathbb{R}^{d_1 \times d_0}\),
\(\vec{w}_2 \in \mathbb{R}^{d_1}\) and elementwise activations
\(\vec{A}_1: \mathbb{R}^{d_1} \to \mathcal{A}_1^{d_1} \subseteq \mathbb{R}^{d_1}\) and
\(A_2: \mathbb{R} \to \mathcal{Y}\). The distribution \(Q(\theta) =: Q_2(\vec{w}_2) Q_1(W_1)\)
factorises over the two layers. This is identical to the above and sets
\(\vec{\eta}_{W_1}(\vec{x}) = \vec{A}_1(W_1 \vec{x})\).

Sampling \(\vec{a}\) is straightforward; one method involves analytically
finding the distribution on the ``pre-activations'', a trivial operation for the
normal distribution among others, before sampling this and passing through the
activation. In combination with the pathwise gradient estimator, this is known
as the ``local reparameterization trick'' \citep{NIPS2015_5666}, and can lead to
considerable computational savings on parallel minibatches compared to direct
hierarchical sampling, \(\vec{a} = A_1(W_1 \vec{x})\) with \(W_{1} \sim Q_{1}\). We
will utilise this in all our reparameterizable dense networks, and a variation
on it in \Cref{all_sign}.

%% file: our_files/section_sign_output.tex
\section{Aggregating Signed-Output Neural Networks}\label{sign_output}

Here we consider multi-layer feed-forward networks with a final sign activation
function and unit variance normal distribution for the final layer,
\(Q^{\vec{w}}(\vec{w}) = \mathcal{N}(\vec{\mu}, \mathbb{I})\), so the
aggregate is given by \eqref{erf_aggregate}. Using \eqref{basic_monte_carlo} and
\eqref{aggregated_estimator} with independent samples
\(\{(\vec{w}^{t}, \theta^{\notw, (t)})\}_{t=1}^{T} \sim Q\) and
\(\vec{\eta}^t := \vec{\eta}_{\theta^{\notw, (t)}}(\vec{x})\) leads to the two
estimators\footnote{Henceforth assuming the technical condition
\(\mathbb{P}_{\vec{\eta}|\vec{x}} \{\vec{\eta} = \vec{0}\} = 0\) that allows aggregation to be
well-defined.}

\begin{align}\label{sign_estimators}
\hat{F}_Q(\vec{x}) &:= \frac{1}{T} \sum_{t=1}^T \sign(\vec{w}^{t} \cdot \vec{\eta}^{t}) &
\hat{F}_Q^{*}(\vec{x}) &:= \frac{1}{T} \sum_{t=1}^T \erf\left(\frac{\vec{\mu} \cdot \vec{\eta}^{t}}{\sqrt{2}\|\vec{\eta}^{t}\|}\right).
\end{align}

Below we discuss how the second estimator, using aggregation of the final layer,
leads to better-behaved training objectives and lower-variance gradient
estimates, enabling stable training of a previously difficult network type.

\subsection{Lower Variance Estimates of Aggregated Sign-Output Networks}

\begin{proposition}\label{aggregated_variance}
With the definitions given in \eqref{sign_estimators}, for all
\(\vec{x} \in \mathbb{R}^{d_0}\), \(T \in \mathbb{N}\), and \(Q\) with normally-distributed final layer,
\[ \mathbb{V}_Q[\hat{F}_Q^{*}(\vec{x})] \leq \mathbb{V}_Q[\hat{F}_Q(\vec{x})]. \]
\end{proposition}

\begin{proof}
Treating \(\vec{\eta}\) as a random variable, always conditioned on \(\vec{x}\),

\begin{align*}
\mathbb{V}_Q[\hat{F}_Q(\vec{x})] &= \frac{1}{T} \mathbb{V}[\sign(\vec{w} \cdot \vec{\eta})] = \frac{1}{T}( 1 - |F_Q(\vec{x})|^{2}) \\
\mathbb{V}_Q[\hat{F}_Q^{*}(\vec{x})] &= \frac{1}{T} \mathbb{V}\left[ \erf \left(\frac{\vec{\mu} \cdot \vec{\eta}}{\sqrt{2}\|\vec{\eta}\|} \right) \right]
                      = \frac{1}{T} \left( \mathbb{E}_{\vec{\eta}}\left| \erf\left(\frac{\vec{\mu} \cdot \vec{\eta}}{\sqrt{2}\|\vec{\eta}\|}\right) \right|^{2} - |F_Q(\vec{x})|^{2} \right). \\
\end{align*}

The result follows as the error function is point-wise smaller than 1.
\end{proof}

\begin{proposition}
Under the conditions of \Cref{aggregated_variance} and \(y \in \{+1, -1\}\),
\[ \mathbb{V}_Q[\ell_{\mathrm{lin}}(\hat{F}_Q^{*}(\vec{x}), y)] \leq \mathbb{V}_Q[\ell_{\mathrm{lin}}(\hat{F}_Q(\vec{x}), y)] \leq \mathbb{V}_{f \sim Q}[\ell_{0-1}(f(\vec{x}), y)] = \frac{1}{4}(1 - |F_Q(\vec{x})|^2).\]
\end{proposition}

\begin{proof}
\[ \mathbb{V}_Q[\ell_{\mathrm{lin}}(\hat{F}_Q(\vec{x}), y)] = \mathbb{E}_Q \left|\frac{1}{2}(yF_Q(\vec{x}) - y\hat{F}_Q(\vec{x}))\right|^2 = \frac{1}{4} \mathbb{V}_Q[\hat{F}_Q(\vec{x})] \]
and a similar result for \(\hat{F}^{*}_Q\). \(f = \hat{F}_Q\) if \(T = 1\) and
\(\ell_{\text{lin}}(f(\vec{x}), y) = \ell_{0-1}(f(\vec{x}), y)\). The result then
follows from this and \Cref{aggregated_variance}.
\end{proof}

Note that as \(F_Q(\vec{x}) \to \pm 1\), both variances disappear, and that the
difference in variances of \(\hat{F}_Q^{*}\) and \(\hat{F}_Q\) will be biggest in
the regime \(\|\vec{\mu}\| \ll 1\).

\subsection{Lower Variance Gradient Estimators}

Final layer derivative estimators given through REINFORCE and aggregation are
derived from the two restatements below:
\begin{align*}
  \vec{G}(\vec{x}) &:= \frac{\partial}{\partial\vec{\mu}} F_Q(\vec{x})
  = \mathbb{E}_{\vec{w}} \mathbb{E}_{\vec{\eta}|\vec{x}} \sign(\vec{w} \cdot \vec{\eta}) (\vec{\mu} - \vec{w})
  = \mathbb{E}_{\vec{\eta}|\vec{x}} \frac{\vec{\eta}}{\|\vec{\eta}\|} \sqrt{\frac{2}{\pi}} \exp\left[ - \frac{1}{2} \left( \frac{\vec{\mu} \cdot \vec{\eta}}{\|\vec{\eta}\|} \right)^2 \right].
\end{align*}

This gives rise to the gradient estimators (using the samples from
\eqref{sign_estimators}):
\begin{align*}
  \hat{\vec{G}}(\vec{x}) &:= \frac{1}{T} \sum_{t=1}^{T} \sign(\vec{w}^{t} \cdot \vec{\eta}^{t}) (\vec{\mu} - \vec{w}^{t})  &
  \hat{\vec{G}}^{*}(\vec{x}) &:= \frac{1}{T} \sum_{t=1}^{T} \frac{\vec{\eta}^{t}}{\|\vec{\eta}^{t}\|} \sqrt{\frac{2}{\pi}} \exp\left[ - \frac{1}{2} \left( \frac{\vec{\mu} \cdot \vec{\eta}^{t}}{\|\vec{\eta}^{t}\|} \right)^2 \right]
\end{align*}

\begin{proposition}
Under the conditions of \Cref{aggregated_variance} and with these
definitions,
\[ \mathrm{Cov}[\hat{\vec{G}}^{*}(\vec{x})] \prec \mathrm{Cov}[\hat{\vec{G}}(\vec{x})]\]
where \(A \prec B \iff B - A\) is positive definite. Thus for all \(\vec{u} \neq \vec{0}\),
\(\mathbb{V}[\hat{\vec{G}}^{*}(\vec{x}) \cdot \vec{u}] < \mathbb{V}[\hat{\vec{G}}(\vec{x}) \cdot \vec{u}]\).
\end{proposition}

\begin{proof}
It is straightforward to show that
\begin{align*}
  \text{Cov}[\hat{\vec{G}}(\vec{x})] &= \frac{1}{T} \left( \mathbb{I} - \vec{G}\vec{G}^T \right) &
  \text{Cov}[\hat{\vec{G}}^{*}(\vec{x})] &= \frac{1}{T} \left( \mathbb{E}\left[\frac{\vec{\eta}\vec{\eta}^{T}}{\|\vec{\eta}\|^{2}}  \frac{2}{\pi} e^{-\left(\frac{\vec{\mu} \cdot \vec{\eta}}{\|\vec{\eta}\|} \right)^2}\right] - \vec{G}\vec{G}^T \right)
\end{align*}
so for \(\vec{u} \neq \vec{0}\),
\begin{align*}
  T \vec{u}^{T} \left( \mathrm{Cov}[\hat{\vec{G}}(\vec{x})] - \mathrm{Cov}[\hat{\vec{G}}^{*}(\vec{x})] \right) \vec{u}
  = \|\vec{u}\|^2 - \frac{2}{\pi} \mathbb{E} \left[\frac{|\vec{u} \cdot \vec{\eta}|^2}{\|\vec{\eta}\|^{2}}  e^{-\left(\frac{\vec{\mu} \cdot \vec{\eta}}{\|\vec{\eta}\|} \right)^2} \right]
  \geq \|\vec{u}\|^2 \left(1 - \frac{2}{\pi}\right) > 0.
\end{align*}
The first equality follows directly from the linearity of the expectation.
\end{proof}

If the rest of the layers are reparameterizable, we can also go on to use the
pathwise estimator to estimate gradients in the aggregated case, which is not
possible otherwise. We show experimentally this is significantly easier.
Firstly, though, we consider an important special case where this is not true: a
feed forward network with all sign activations.

\subsection{All Sign Activations}\label{all_sign}

Choosing all sign activations and unit-variance normal distributions on the
weights of each feed-forward layer,

\[ f_\theta(\vec{x}) = \sign(\vec{w}_L \cdot \sign(W_{L-1} \dots \sign(W_1 \vec{x}) \dots )) \]

with \(\theta := \operatorname{vec}(\vec{w}_L, \dots, W_1)\) and
\(W_l := [\vec{w}_{l, 1} \dots \vec{w}_{l, d_l}]^T\); \(l \in \{1, ..., L\}\)
indexes layers. The distribution factorises into
\(Q_l(W_l) = \prod_{i=1}^{d_l} q_{l, i}(\vec{w}_{l, i})\) with
\(q_{l, i} = \mathcal{N}(\vec{\mu}_{l, i}, \, \mathbb{I}_{d_{l-1}})\).

In the notation of Section \ref{aggregates},
\(\vec{\eta}_{\thetanotw}(\vec{x}) = \sign(W_{L-1} \dots \sign(W_1 \vec{x}) \dots )\) is
the final layer activation, which could easily be sampled by mapping \(\vec{x}\)
through the first \(L-1\) layers with draws from the weight distribution.
Instead, we go on to make an iterative replacement of the weight distributions
on each layer by conditionals on the layer activations to obtain the summation

\begin{equation}\label{iterated_integral}
F_Q(\vec{x})
  = \sum_{\vec{a}_1 \in \{+1, -1\}^{d_1}} \dots \sum_{\vec{a}_{L-1} \in \{+1, -1\}^{d_{L-1}}} \erf\left( \frac{\vec{\mu}_{L} \cdot \vec{a}_{L-1}}{\sqrt{2}\|\vec{a}_{L-1}\|} \right) \tilde{Q}_{L-1}(\vec{a}_{L-1}|\vec{a}_{L-2}) \dots \tilde{Q}_1(\vec{a}_1|\vec{x})
\end{equation}

and hierarchically sample the \(\vec{a}_l\); like local-reparameterisation, this
leads to a considerable computational saving over sampling a separate weight
matrix for every input. The conditionals can be found in closed form as
\(\tilde{Q}_l(\vec{a}_l|\vec{a}_{l-1}) := \prod_{i=1}^{d_l} \tilde{q}_{l, i}(a_{l, i} | \vec{a}_{l-1})\),
and (with \(\vec{a}_0 := \vec{x}\))

\[
\tilde{q}_{l, i}(a_{l, i} = \pm 1\, |\vec{a}_{l-1}) = \int_0^\infty \mathcal{N}(\pm\vec{\mu}_{l, i} \cdot \vec{a}_{l-1}, \|\vec{a}_{l-1}\|^2) \D{z} = \frac{1}{2} \left[ 1 \pm \erf \left( \frac{\vec{\mu}_{l, i} \cdot \vec{a}_{l-1}}{\sqrt{2}\|\vec{a}_{l-1}\|} \right) \right].
\]

A marginalised REINFORCE-style gradient estimator for \emph{conditional}
distributions can then be used; this does not necessarily have better
statistical properties but in combination with the above is much more
computationally efficient. Using samples
\(\{(\vec{a}^{t}_1 \dots \vec{a}^{t}_{L-1})\}_{t=1}^T \sim \tilde{Q}\),
\begin{equation}\label{local_reinforce}
\frac{\partial F_Q(\vec{x})}{\partial\vec{\mu}_{l, i}} \approx \frac{1}{T} \sum_{t=1}^T \erf\left( \frac{\vec{\mu}_{L} \cdot \vec{a}^{t}_{L-1}}{\sqrt{2}\|\vec{a}^{t}_{L-1}\|} \right) \, \frac{\partial}{\partial\vec{\mu}_{l, i}} \log\tilde{q}_{l, i}(a^{t}_{l, i} |\vec{a}^{t}_{l-1}).
\end{equation}

The above formulation somewhat resembles the PBGNet model of
\cite{NIPS2019_8911}, but derived in a very different way. Both are equivalent
in the single-hidden-layer case, but with more layers PBGNet uses an unusual
tree-structured network for which the exact aggregate can be calculated (very
expensively, though avoiding an exponential dependency on depth). Generally,
though, to avoid this cost, they resort to a Monte Carlo approximation:
informally, this draws new samples for every layer \(l\) based on an average of
those from the previous layer,
\(\vec{a}_l | \{\vec{a}_{l-1}^{(t)}\}_{t=1}^T \sim \frac{1}{T} \sum_{t=1}^T \tilde{Q}(\vec{a}_l|\vec{a}_{l-1}^{(t)})\).

This is all justified within the tree-structured framework but leads to an
exponential KL penalty which---as hinted by \citet{NIPS2019_8911} and shown
empirically in \Cref{experiments}---makes PAC-Bayes bound optimisation strongly
favour shallower such networks. In addition to this practical drawback, our
formulation is more general and we claim it makes the fundamental ideas of
partial-aggregation and marginalised sampling significantly clearer.

%% file: our_files/section_objective_bound.tex
\section{PAC-Bayesian Objectives with Signed-Outputs}\label{bounds}

We now move to obtain binary classifiers with guarantees for the expected
misclassification error, \(R^{0-1}\), which we do by optimizing PAC-Bayesian
bounds. Such bounds (as in \Cref{catonizhou}) will usually involve the
non-differentiable and non-convex misclassification loss \(\ell_{0-1}\). However,
to train a neural network we need to replace this by a differentiable surrogate,
as discussed in the introduction.

Here we adopt a different approach by using our signed-output networks, where
since \(f(\vec{x}) \in \{+1, -1\}\), there is an exact equivalence between the
linear and misclassification losses,
\(\ell_{0-1}(f(x), y) = \ell_{\text{lin}}(f(x), y)\), avoiding the extra factor of two
from the inequality \(\ell_{0-1} \leq 2\ell_{\text{lin}}\) used by \citet[][Section
2.1]{NIPS2019_8911}.

Although we have only moved the non-differentiability into \(f\), the form of a
PAC-Bayesian bound and the linearity of the loss and expectation allow us to go
further and aggregate,

\begin{equation} \label{linear01}
\mathbb{E}_{f \sim Q} \ell_{0-1}(f(\vec{x}), y') = \mathbb{E}_{f \sim Q} \ell_{\text{lin}}(f(x), y) = \ell_{\text{lin}}(F_Q(\vec{x}), y')
\end{equation}

which as we saw in \Cref{aggregated_signed_linear} can make some such
non-differentiable functions differentiable.

Combining \eqref{linear01} with Theorem \ref{catonizhou}, we obtain a directly
optimizable, differentiable bound on the misclassification loss without
introducing an extra factor of 2:

\begin{theorem}
Given \(P\) on \(\theta\) and \(\alpha > 1\), for all \(Q\) on \(\theta\) and \(\lambda > 1\)
simultaneously with probability at least \(1 - \delta\) over \(S \sim \mathcal{D}^m\),
\[ \mathbb{E}_{\theta \sim Q}R^{0-1}(f_\theta) \leq \Phi_{\lambda/m}^{-1} \left[ R^{\mathrm{lin}}_S(F_Q) + \frac{\alpha}{\lambda} \left( \KL{Q|P} - \log \delta + 2\log \left(\frac{\log \alpha^{2}\lambda}{\log{\alpha}}\right) \right) \right] \]
with \(\Phi_\gamma^{-1}(t) = \frac{1 - \exp(-\gamma t)}{1 - \exp(-\gamma)}\) and \(f_\theta: \mathbb{R}^{d} \to \{+1, -1\}, \, \theta \in \theta\).\label{main_bound}
\end{theorem}

Thus, for each \(\lambda\), which can be held fixed (``\textbf{fix-\(\lambda\)}'') or
simultaneously optimized throughout training for automatic regularisation tuning
(``\textbf{optim-\(\lambda\)}''), we obtain a gradient descent objective:

\begin{equation}\label{basic_objective}
R^{\text{lin}}_S (\hat{F}^{*}_Q) + \frac{\text{KL}(Q|P)}{\lambda}.
\end{equation}

%% file: our_files/section_experiments.tex
\section{Experiments}\label{experiments}

All experiments run on ``binary''-MNIST, dividing MNIST into two classes, of
digits 0-4 and 5-9. Neural networks had three hidden layers with 100 units per
layer and \textbf{sign}, sigmoid (\textbf{sgmd}) or \textbf{relu} activations,
before a single-unit final layer with sign activation. \(Q\) was chosen as an
isotropic, unit-variance normal distribution with initial means drawn from a
truncated normal distribution of variance \(0.05\). The data-free prior \(P\)
was fixed equal to the initial \(Q\), as motivated by \citet[][Section 5 and
Appendix B]{dziugaite2017computing}.

The objectives \textbf{fix-\(\lambda\)} and \textbf{optim-\(\lambda\)} from \Cref{bounds}
were used for batch-size 256 gradient descent with Adam \citep{kingma2014adam}
for 200 epochs. Every five epochs, the bound (for a minimising \(\lambda\)) was
evaluated using the entire training set; the learning rate was then halved if
the bound was unimproved from the previous two evaluations. The best
hyperparameters were selected using the best bound achieved in these evaluations
through a grid search of initial learning rates \(\in \{0.1, 0.01, 0.001\}\),
sample sizes \(T \in \{1, 10, 50, 100\}\). Once these were selected training was
repeated 10 times to obtain the values in \Cref{mainresults}.

\(\lambda\) in \textbf{optim-\(\lambda\)} was optimised through \Cref{main_bound} on
alternate mini-batches with SGD and a fixed learning rate of \(10^{-4}\) (whilst
still using the objective \eqref{basic_objective} to avoid effectively scaling
the learning rate with respect to empirical loss by the varying \(\lambda\)). After
preliminary experiments in \textbf{fix-\(\lambda\)}, we set \(\lambda = m = 60000\), the
training set size, as is common in Bayesian deep learning.

We also report the values of three baselines: \textbf{reinforce}, which uses the
fix-\(\lambda\) objective without partial-aggregation, forcing the use of REINFORCE
gradients everywhere; \textbf{mlp}, an unregularised non-stochastic relu neural
network with \(\tanh\) output activation; and the PBGNet model (\textbf{pbg})
from \citet{NIPS2019_8911}, with the misclassification error bound obtained
through \(\ell_{0-1} \leq 2\ell_{\text{lin}}\). Despite significant additional
hyperparameter exploration for the latter, we were unable to train a three layer
network through the PBGNet algorithm directly comparable to our method, likely
because of the exponential KL penalty (in their equation 17) within that
framework; to enable comparison, we therefore allowed the number of hidden
layers in this scenario to vary \(\in \{1, 2, 3\}\). Other baseline tuning and
setup was similar to the above, see the Appendix for more details.

\begin{table}[]
  \caption{Average (from ten runs) binary-MNIST losses and bounds (\(\delta=0.05\))
for the best epoch and optimal hyperparameter settings of various algorithms.
Hyperparameters and epochs were chosen by bound if available and non-vacuous,
otherwise by training linear loss.} \centering

\begin{tabular}{@{}ccccccccccc@{}}
\toprule
                                          & \textbf{mlp}  & \textbf{pbg}  & \multicolumn{2}{c}{\textbf{reinforce}} & \multicolumn{3}{c}{\textbf{fix-\(\lambda\)}}        & \multicolumn{3}{c}{\textbf{optim-\(\lambda\)}}      \\ \midrule
                                          &               &               & sign               & relu              & sign          & sgmd       & relu          & sign          & sgmd       & relu          \\ \midrule
\multicolumn{1}{l}{\textbf{Train Linear}} & 0.78          & 8.72          & 26.0               & 18.6              & 8.77          & 7.60          & 6.35          & 6.71          & 6.47          & 5.41          \\
\textit{error, \(1\sigma\)}               & \textit{0.08} & \textit{0.08} & \textit{0.8}       & \textit{1.4}      & \textit{0.04} & \textit{0.19} & \textit{0.10} & \textit{0.11} & \textit{0.18} & \textit{0.16} \\
\multicolumn{1}{l}{\textbf{Test 0-1}}     & 1.82          & 5.26          & 25.4               & 17.9              & 8.73          & 7.88          & 6.51          & 6.85          & 6.84          & 5.61          \\
\textit{error, \(1\sigma\)}               & \textit{0.16} & \textit{0.18} & \textit{1.0}       & \textit{1.5}      & \textit{0.23} & \textit{0.30} & \textit{0.19} & \textit{0.27} & \textit{0.21} & \textit{0.20} \\
\multicolumn{1}{l}{\textbf{Bound 0-1}}    & -             & 40.8          & 100                & 100               & 21.7          & 18.8          & 15.5          & 22.6          & 19.3          & 16.0          \\
\textit{error, \(1\sigma\)}               & \textit{-}    & \textit{0.2}  & \textit{0.0}       & \textit{0.0}      & \textit{0.04} & \textit{0.17} & \textit{0.04} & \textit{0.03} & \textit{0.31} & \textit{0.05} \\ \bottomrule
\end{tabular}

\label{mainresults}
\end{table}

%% file: our_files/section_discussion.tex
\section{Discussion}\label{discussion}

The experiments demonstrate that partial-aggregation enables training of
multi-layer non-differentiable neural networks in a PAC-Bayesian context:
REINFORCE gradients and a multiple-hidden-layer PBGNet \citep{NIPS2019_8911}
obtained only non-vacuous bounds, and our misclassification bounds improve those
of a single-hidden-layer PBGNet. We note that our bound optimisation is
empirically quite conservative, and the non-stochastic mlp model obtains a lower
overall error; understanding this gap is one of the key questions in the theory
of deep learning. Finally, we also observe that using
\(\sign(\hat{F}_Q(\vec{x}))\) with \(T > 1\) for test prediction, as in PBGNet,
gave improved empirical results despite the inferior theoretical guarantees; we
consider this an interesting avenue of future research.

%% file: our_files/appendix_experiments.tex
\section{Further Experimental Details}\label{extra_experiments}

\subsection{Aggregating Biases with the Sign Function}

We used a bias term in our network layers, leading to a simple extension of the
above formulation, omitted in the main text for conciseness:
\[ E_{\vec{w} \sim \mathcal{N}(\vec{\mu}, \Sigma), b \sim \mathcal{N}(\beta, \sigma^2)} \, \sign(\vec{w} \cdot \vec{x} + b) = \erf\left(\frac{\vec{\mu} \cdot \vec{x} + \beta}{\sqrt{2(\vec{x}^T\Sigma\vec{x} + \sigma^2)}}\right)\]
since \(\vec{w} \cdot \vec{x} + b \sim \mathcal{N}(\vec{\mu} \cdot \vec{x} + \beta, \vec{x}^T\Sigma\vec{x} + \sigma^2)\) and
\begin{align*}
  E_{z \sim \mathcal{N}(\alpha, \beta^2)} \sign z &= P(z \geq 0) - P(z < 0) \\
  &= [1 - \Phi(-\alpha/\beta)] - \Phi(-\alpha/\beta) \\
  &= 2\Phi(\alpha/\beta) - 1 = \erf(\alpha/\sqrt{2}\beta). \\
\end{align*}

The bias and weight co-variances were chosen to be diagonal with a scale of 1,
which leads to some simplification in the above.

\subsection{Reinforce Model}

During evaluation, the \textbf{reinforce}, draws a new set of weights for every
test example, equivalent to the evaluation of the other models; but doing so
during training, with multiple parallel samples, is prohibitively expensive.

Two different approaches to straightforward, not partially-aggregated, gradient estimation for the baseline \textbf{reinforce} suggest themselves, arising from different approximations to the \(Q\)-expected loss of the minibatch, \(B \subseteq S\) (with data indices \(\mathcal{B}\)). From the identities

\begin{equation*}
\nabla_\phi \E_{\theta \sim q_\phi} R_B(f_\theta)
= \E_{\theta \sim q_\phi}  \frac{1}{|B|} \sum_{i\in\mathcal{B}} \ell(f_\theta(\vec{x}_i), y_i) \nabla_\phi \log q_\phi(\theta)
= \frac{1}{|B|} \sum_{i\in\mathcal{B}} \E_{\theta \sim q_\phi}  \ell(f_\theta(\vec{x}_i), y_i) \nabla_\phi \log q_\phi(\theta)
\end{equation*}

we obtain two slightly different estimators for \(\nabla_\phi \E_{\theta \sim q_\phi} R_B(f_\theta)\):
\begin{align*}
&\frac{1}{T|B|} \sum_{t=1}^T \sum_{i\in\mathcal{B}} \ell(f_{\theta^{(t, i)}}(\vec{x}_i), y_i) \nabla_\phi \log q_\phi(\theta^{(t, i)})      &
&\frac{1}{T|B|} \sum_{i\in\mathcal{B}} \sum_{t=1}^T \ell(f_{\theta^t}(\vec{x}_i), y_i) \nabla_\phi \log q_\phi(\theta^t)
\end{align*}

The first draws many more samples and has lower variance but is much slower
computationally; even aside from the \(O(|B|)\) increase in computation, there
is a slowdown as the optimised BLAS matrix routines cannot be used, and the very
large matrices involved may not fit in memory \cite[see][for more
information]{NIPS2015_5666}.

Therefore, as is standard in the Bayesian Neural Network literature with the
pathwise estimator, we use the latter formulation, which has a similar
computational complexity to local-reparameterisation and our marginalised
REINFORCE estimator \eqref{local_reinforce}. We should note though that in
preliminary experiments, the alternate estimator did not appear to lead to
improved results. This clarifies the advantages of marginalised sampling, which
can lead to lower variance with a similar computational cost.

\subsection{Dataset Details}

We used the MNIST dataset version 3.0.1, available online at
\url{http://yann.lecun.com/exdb/mnist/}, which contains 60000 training examples
and 10000 test examples, which were used without any further split, and rescaled
to lie in the range \([0, 1]\). For the ``binary''-MINST task, the labels \(+1\)
and \(-1\) were assigned to digits in \(\{5, 6, 7, 8, 9\}\) and
\(\{0, 1, 2, 3, 4\}\) respectively, and images were scaled into the interval
\([0, 1]\).

\subsection{Hyperparameter Search for Baselines}\label{baseline_search}

The baseline comparison values offered with our experiments were optimized
similarly to the above, for completeness we report everything here.

The MLP model had three hidden ReLu layers of size 100 each trained with Adam, a
learning rate \(\in \{0.1, 0.01, 0.001\}\) and a batch size of 256 for 100 epochs.
Complete test and train evaluation was performed after every epoch, and in the
absence of a bound, the model and epoch with lowest train linear loss was
selected.

For PBGNet we choose the values of hyperparameters from within these values
giving the least bound value. Note that, unlike in the original paper, we do not
allow the hidden size to vary \(\{\in {10, 50, 100}\}\), and we use the entire
MNIST training set as we do not need a validation set. While attempting to train
a three hidden layer network, we also searched through the hyperparameter
settings with a batch size of \(64\) as in the original, but after this failed,
we returned to the original batch size of \(256\) with Adam. All experiments were
performed using the code from the original paper, available at
\url{https://github.com/gletarte/dichotomize-and-generalize}.

Since we were unable to train a multiple-hidden-layer network through the PBGNet
algorithm, for this model only we explored different numbers of hidden layers
\(\in \{1, 2, 3\}\).

\subsection{Final Hyperparameter Settings}

In \Cref{hparam_settings} we report the hyperparameter settings used for the
experiments in \Cref{mainresults} after exploration. To save computation,
hyperparameter settings that were not learning (defined as having a
whole-train-set linear loss of \(> 0.45\) after ten epochs) were terminated
early. This was also done on the later evaluation runs, where in a few instances
the fix-\(\lambda\) sigmoid network failed to train after ten epochs; to handle this
we reset the network to obtain the main experimental results.

For clarity we repeat here the hyperparameter settings and search space:

\begin{itemize}
\item Initial Learning Rate \(\in \{0.1, 0.01, 0.001\}\).
\item Training Samples \(\in \{1, 10, 50, 100\}\).
\item Hidden Size \(= 100\).
\item Batch Size \(= 256\).
\item Fix-\(\lambda\), \(\lambda = m = 60000\).
\item Number of Hidden Layers \(= 3\) for all models, except PBGNet \(\in \{1, 2, 3\}\).
\end{itemize}

\begin{table}[]
  \caption{Chosen Hyperparameter settings and additional details for results in
\Cref{mainresults}. Best hyperparameters were chosen by bound if available and
non-vacuous, otherwise by best training linear loss through a grid search as
described in \Cref{experiments} and \Cref{baseline_search}. Run times are
rounded to nearest 5 minutes.} \centering

\begin{tabular}{lcccccccccc}
\hline
            & \textbf{mlp} & \textbf{pbg} & \multicolumn{2}{c}{\textbf{reinforce}} & \multicolumn{3}{c}{\textbf{fix-\(\lambda\)}} & \multicolumn{3}{c}{\textbf{optim-\(\lambda\)}} \\ \hline
            &              &              & sign               & relu              & sign        & relu        & sgmd       & sign         & relu        & sgmd        \\ \hline
Init. LR    & 0.001        & 0.01         & 0.1                & 0.1               & 0.01        & 0.1         & 0.1        & 0.01         & 0.1         & 0.1         \\
Samples, T  & -            & 100          & 100                & 100               & 100         & 50          & 10         & 100          & 100         & 10          \\
Hid. Layers & 3            & 1            & 3                  & 3                 & 3           & 3           & 3          & 3            & 3           & 3           \\
Hid. Size   & 100          & 100          & 100                & 100               & 100         & 100         & 100        & 100          & 100         & 100         \\
Mean KL     & -            & 2658         & 15020              & 13613             & 2363        & 3571        & 3011       & 5561         & 3204        & 4000        \\
Runtime/min & 10           & 5            & 40                 & 40                & 35          & 30          & 25         & 35           & 30          & 25          \\ \hline
\end{tabular}

\label{hparam_settings}
\end{table}

\subsection{Implementation and Runtime}

Experiments were implemented using Python and the TensorFlow library
\citep{tensorflow2015-whitepaper}.
Reported
approximate runtimes are for execution on a NVIDIA GeForce RTX 2080 Ti GPU.